\documentclass[12pt]{article}
\usepackage{amsthm}
\usepackage{amsmath}
\usepackage{amsfonts}
\usepackage{algpseudocode}
\usepackage{algorithmicx}
\usepackage{algorithm}
\usepackage{multicol}
\usepackage{graphicx}
\usepackage{subfigure}
\usepackage{verbatim}
\usepackage{color}
\usepackage{hyperref}
\newtheorem{proposition}{Proposition}%[section]
\usepackage{pifont}
\renewcommand{\checkmark}{\ding{52}}
\newcommand{\cross}{\ding{55}}
\newcommand{\equal}{\textbf{=}}
\theoremstyle{definition}
\newtheorem{definition}{Definition}%[section]
\theoremstyle{remark}
\usepackage[utf8]{inputenc}

\title{Bayesian Conditional Gaussian Network Classifiers with Applications to Mass Spectra Classification}

\author{Victor Bellón\\
IIIA-CSIC, Spain\\
bellon@iiia.csic.es
\and
Jesús Cerquides\\
IIIA-CSIC, Spain\\
cerquide@iiia.csic.es
\and
Ivo Grosse\\
Martin-Luther-Universität, Germany\\
grosse@informatik.uni-halle.de}

\begin{document}
\maketitle
\begin{abstract}
Classifiers based on probabilistic graphical models %\cite{Lauritzen1996} 
are very effective. In continuous domains, maximum likelihood is usually used to assess the predictions of those classifiers. When data is scarce, this can easily lead to overfitting. In any probabilistic setting, Bayesian averaging (BA) provides theoretically optimal predictions and is known to be robust to overfitting. In this work we introduce Bayesian Conditional Gaussian Network Classifiers, which efficiently perform exact Bayesian averaging over the parameters. We evaluate the proposed classifiers against the maximum likelihood alternatives proposed so far over standard UCI datasets, concluding that performing BA improves the quality of the assessed probabilities (conditional log likelihood) whilst maintaining the error rate. 

Overfitting is more likely to occur in domains where the number of data items is small and the number of variables is large. These two conditions are met in the realm of bioinformatics, where the early diagnosis of cancer from mass spectra is a relevant task. We provide an application of our classification framework to that problem, comparing it with the standard maximum likelihood alternative, where the improvement of quality in the assessed probabilities is confirmed.
\end{abstract}

%\end{frontmatter}

\section{Introduction}

%\subsection{Probabilisitic classification}
% \subsection{Chain graphs}
% A class of graphs of special interest to us is the class of \emph{chain graphs}. These are the graphs where the vertex set $V$ can be partitioned into numbered subsets, forming a so-called \emph{dependence chain} $V=V(1)\cup\ldots\cup V(T)$
% such that all edges between vertices in the same subset are undirected and all edges between different subsets are directed, pointing from the set with lower number to the one with higher number. 

Supervised classification is a basic task in data analysis and pattern recognition. It requires the construction of a classifier, i.e. a function that assigns a class label to instances described by a set of variables. There are numerous classifier paradigms, among which the ones based on probabilistic graphical models (PGMs) \cite{Lauritzen1996}, are very effective and well-known in domains with uncertainty.

A widely used assumption is that data follows a multidimensional Gaussian distribution\cite{Geiger1994}. This is adapted for classification problems by assuming that data follows a multidimensional Gaussian distribution that is different for each class, encoding the resulting distribution as a Conditional Gaussian Network (CGN)\cite{Bottcher2004}. In \cite{Larranaga2006}, Larrañaga, Pérez and Inza introduce and evaluate classifiers based on CGNs with a more detailed description in \cite{Perez2010}. They analyze different methods to identify a Bayesian network structure and a set of parameters such that the resultant CGN performs well in the classification task.  In \cite{Perez2010} the same authors propose to estimate the parameters directly from the sample mean and sample covariance matrix in the data, that is, using maximum likelihood (ML). Following this strategy can lead to model overfitting when data is scarce. In bioinformatics, models are sought in domains where the number of data items is small and the number of variables is large, such as classification of mass spectrograms or microarrays. To try to avoid overfitting, we propose classifiers based on CGNs that instead of estimating the parameters by ML, perform exact Bayesian averaging over them, and we conclude that they provide more accurate estimates of probabilities without sacrificing accuracy.

We start the paper by introducing Bayesian networks and reviewing their use for classification in section~\ref{sec:BNCs}. After that, we define CGNs formally in section~\ref{sec:CGNs}. Then, in section~\ref{sec:ML}, we review the theoretical results from \cite{Lauritzen1996} that provide the foundation to assess parameters in CGNs using the ML principle. In section~\ref{sec:BMA} we review and state in a more formal way some results appearing in \cite{Bottcher2004} for averaging over parameters in CGNs. In section~\ref{sec:Experimental} we compare the results of both strategies over UCI datasets and in section~\ref{sec:Cancer} we compare them for the case of early diagnosis of ovarian cancer from mass spectra. We conclude providing future research lines in section~\ref{sec:Conclusions}.

The main contribution of the paper is noticing that, (i) current state-of-the-art work in CGN classifiers \cite{Perez2010,Larranaga2006}, disregard the possibility of performing Bayesian averaging, and that (ii) the quality of the estimated probability significantly improves if we use it. Thus, we restate the results of \cite{Bottcher2004} for the specific case of classification and with a clear algorithmic perspective, so that they can be easily applied by other researchers interested in reaping the benefits of Bayesian averaging in CGN classifiers.

\section{Bayesian network classifiers}
\label{sec:BNCs}
In this section we introduce the notation to be used in the paper, discuss what Bayesian networks are, and review different approximations to learn classifiers based on Bayesian networks in the literature. 

\subsection{Notation}
\label{sec:Notation}
The notation used in the paper is very similar to the one used by Bøttcher and Lauritzen in \cite{Lauritzen1996,Bottcher2004}. Let $X$ be a set of random variables used to des\-cribe an object. We define a set of indexes $V$, one for each variable, that is, $X=(x_v)_{v\in V}$. In this paper, we deal with two different types of random variables: discrete and continuous. We use $I$ and $Y$ to refer to the set of discrete and continuous random variables respectively. We assume that the set of indexes $V=\Delta \cup \Gamma$, where $\Delta$ and $\Gamma$ are the disjoint sets of discrete and continuous variable indexes respectively. That is: $X=(I,Y)=((I_\delta)_{\delta \in \Delta},(Y_\gamma)_{\gamma \in \Gamma}).$ Each discrete variable $I_\delta$ takes values over the finite\footnote{Infinite discrete random variables are not considered in this work} set ${\mathcal I}_\delta$  and each continuous variable takes values over $\mathbb R.$ In the following, $\mathbf{x}$ represents an assignment of values to the variables in $X$. Furthermore,  $\mathbf{x}=(\mathbf{i},\mathbf{y})$, where  $\mathbf{i}$ is an assignment of values to the variables in $I$, and $\mathbf{y}$ is an assignment of values to the variables in 
$Y$. Given a set of indexes $A$, $\mathbf{i}_A$ (resp. $\mathbf{y}_A$, $\mathbf{x}_A$) represents the restriction of $\mathbf{i}$ (resp. $\mathbf{y}$, $\mathbf{x}$) to variables with index in $A\cap \Delta$ (resp. $A\cap \Gamma$, $A$). 

\subsection{Bayesian networks}
A Bayesian network is a probabilistic graphical model \cite{Lauritzen1996,Koller2009} that encodes the joint probability distribution for a set of random variables $X$. A directed acyclic graph (DAG) $D=(V,E)$, where $V$ is the set of vertexes and $E$ is the set of edges, encodes the \emph{structure} of the Bayesian network. Each vertex $v \in V$ is associated with a random variable $X_v$. Let $pa(v)$ be the set of parents of $v$ in $D$. To each vertex $v$, we attach a probability distribution $p(x_v|\mathbf{x}_{pa(v)})$. The probability distribution encoded in the Bayesian network is 
\begin{equation}
 p(\mathbf{x})=\prod_{v\in V}p(x_v|\mathbf{x}_{pa(v)})
\end{equation}

Usually, the probability distribution for each vertex $v$ is part of a parametric family, that is, depends exclusively on a set of parameters associated to vertex $v$, which we will note $\theta_v$. The set $\Theta_V = \{\theta_v\ |\ v\in V\}$ contains the \emph{parameters} of the Bayesian network, while $D$ is its \emph{structure}.

Many works in Bayesian networks make the assumption that data contains only discrete variables. Two different alternatives are usually considered in the literature when data contains both discrete and continuous variables. Eventually, continuous variables can be discretized so as to use discrete Bayesian network classifiers. Alternatively, continuous variables can be directly modeled. This is usually done by assuming that the conditional distribution of a continuous variable given their parents belongs to a parametric family. The most widely used distributional assumption is assuming \emph{conditional Gaussianity}. Bayesian networks making this assumption are known as conditional Gaussian networks (CGN) and are the models that will be studied in this work.

\subsection{Bayesian network classifiers}
The task of classification consists in assigning an input value to one class of a given set of classes. For example, determine whether a picture should be classified as \emph{landscape} or \emph{non-landscape}. Constructing a classifier to produce a posterior probability  $p(class|input)$  is very useful in practical recognition situations, where it allows to take decisions based on a utility model\cite{Duda2000}. Bayesian networks have been successfully used to construct classifiers \cite{Perez2010,Friedman1997,Cerquides2005}. 

Several strategies are possible to apply Bayesian networks for classification. These strategies differ on how we deal with structures and with parameters. 

% Provided the structure is fixed, the simplest alternative is to take a point estimate of its parameters. This is done for instance by the naive Bayes classifier and by logistic regression \cite{Ng2001}. The point estimate is provided by ML for naive Bayes and by maximum conditional likelihood (MCL) for logistic regression. Another possibility is to perform Bayesian learning over the parameters. Kontkanen et al. \cite{Kontkanen1998} showconti how this can be done for the naive Bayes structure. 

When several structures are possible, the simplest alternative is to select a single one and then apply any of the strategies of the previous paragraph to deal with the parameter learning. An example, when we restrict structures to trees, is the Tree Augmented Naive Bayes classifier \cite{Friedman1997}. However, we can also perform Bayesian learning simultaneously over both structures and parameters as is done in \cite{Dash2004,Cerquides2005}. Table~\ref{tab:LearningStrategies} shows some examples of the alternatives for Bayesian network classifiers.

\begin{table}
\begin{small}
\begin{center}
\begin{tabular}{c|c|c}
 \bf{Classifier}              & \bf{Structure} & \bf{Parameters}\\ \hline
 NB \cite{Ng2001}         &  Fixed    & Point estimate by ML \\
 LR \cite{Ng2001}         &  Fixed    & Point estimate by MCL\\
 BIBL \cite{Kontkanen1998}    &  Fixed    & Bayesian \\
 TAN \cite{Friedman1997}      &  Point estimate by ML among trees & Point estimate by ML\\
 NMA \cite{Dash2004}          &  Bayesian among Selective NB & Bayesian \\
 TBMATAN \cite{Cerquides2005} &  Bayesian among trees & Bayesian \\
\end{tabular}
\end{center}
\end{small}
\caption{\label{tab:LearningStrategies}Different Bayesian network classifiers according to their strategies to learn structure and parameters.}
\end{table}

For datasets with mixed variables (both discrete and continuous), CGN classifiers have been proposed in \cite{Perez2006,Perez2010}. There, several different heuristic procedures are proposed to select a single classifier structure. Then, a point estimate for the parameters is provided using ML. In this paper we propose to perform exact Bayesian learning over the parameters in conditional Gaussian network (CGN) classifiers, making use of the results of Bøttcher in \cite{Bottcher2004}. Bayesian learning is the best founded alternative to fit a model to data from the point of view of probability theory. Furthermore, as argued in \cite{Gelman2004}, ``\emph{the central feature of Bayesian inference, the direct quantification of uncertainty, means that there is no impediment in principle to fit problems with many parameters''.} The objective of this paper is to show that the theoretical results in  \cite{Bottcher2004}, allow for a rigurous theoretical treatment of the parameter learning process in CGN 
classifiers. As a result of that, CGN classifiers that use Bayesian learning over parameters provide:
\begin{itemize}
 \item Equivalent results in terms of accuracy (0-1 loss).
 \item Significantly more accurate results in the quality of the probabilities (measured by the average of the CLL of the correct class).
 \item More flexible modeling, provided that we can incorporate prior knowledge into the classification process by means of the prior distribution assumed over parameters.
\end{itemize}
We start by formally introducing the conditional Gaussian network model.
\section{\label{sec:CGNs}Conditional Gaussian networks}
%
%Following \cite{Bottcher2004}, we deal with two different types of random variables: discrete and continuous. Thus, we assume that the set of vertexes $V=\Delta \cup \Gamma$, where $\Delta$ and $\Gamma$ are the sets of discrete and continuous vertexes respectively. We use $I$ and $Y$ to refer to the set of discrete and continuous random variables respectively. That is: $X=(I,Y)=((I_\delta)_{\delta \in \Delta},(Y_\gamma)_{\gamma \in \Gamma}).$ Each discrete variable $I_\delta$ takes values over the set ${\mathcal I}_\delta.$

%As stated in section~\ref{sec:Notation}, our interest is to learn to classify with both discrete and continuous attributes.

Conditional Gaussian networks (CGNs) allow for efficient representation, inference and learning in Bayesian networks that have both discrete and continuous variables. Given a variable index $v$, let $pd(v)=pa(v) \cap \Delta$ be the set of discrete parents of $v$, and $pc(v) = pa(v)\cap \Gamma$ be the set of continuous parents of $v$. In a CGN, discrete variables are not allowed to have continuous parents. That is, for each $\delta \in \Delta$ we have that $pc(\delta)=\emptyset.$ As a consequence, the joint probability distribution factorizes as
\begin{equation}
\label{eq:CGNFactorization}
p(\mathbf{x}) = p(\mathbf{i},\mathbf{y}) = p(\mathbf{i})p(\mathbf{y}|\mathbf{i}) = \prod_{\delta \in \Delta}p(i_\delta|\mathbf{i}_{pa(\delta)})\times \prod_{\gamma\in \Gamma}p(y_\gamma|\mathbf{y}_{pc(\gamma)},\mathbf{i}_{pd(\gamma)})
\end{equation}
where  $\mathbf{i}_{pa(\delta)}$ (resp. $\mathbf{i}_{pd(\gamma)}$) denotes the values of the discrete random variables which are parents of $\delta$ (resp. $\gamma$ ), and $\mathbf{y}_{pc(\gamma)}$ denotes the values of the continuous random variables which are parents of $\gamma$.

Furthermore, in a CGN, the local probability distributions are restricted to conditional multinomial distributions for the discrete nodes and conditional Gaussian linear regressions for the continuous nodes. In the following we provide a parameterization of a CGN. 

\subsection{Distribution over discrete variables}
For each discrete variable index $\delta \in \Delta$ and for each cell of its parents ($\mathbf{i}_{pa(\delta)} \in {\mathcal I}_{pa(\delta)}$), its conditional distribution  $p(i_\delta|\mathbf{i}_{pa(\delta)})$ follows a multinomial distribution\footnote{A reference of the distributions used in the paper can be found at \ref{sec:Distributions}}. We can parameterize it by a vector $\boldsymbol{\theta}_{\delta|\mathbf{i}_{pa(\delta)}}=\{\theta_{i_\delta|\mathbf{i}_{pa(\delta)}} | i_\delta \in {\mathcal I}_\delta\}$ such that  
\begin{alignat}
\theta \theta_{i_\delta|\mathbf{i}_{pa(\delta)}} >0 &  \ \ \ \forall i_\delta \in {\mathcal I}_\delta\\
\sum_{i_\delta\in {\mathcal I_\delta}}\theta_{i_\delta|\mathbf{i}_{pa(\delta)}} = 1 & 
\end{alignat}
Thus, the joint distribution over discrete variables can be parameterized by the set 
\begin{equation}
 \Theta_\Delta=\{\boldsymbol{\theta}_{\delta|\mathbf{i}_{pa(\delta)}} | \ \ \delta \in \Delta;\ \ \  \mathbf{i}_{pa(\delta)} \in {\mathcal I}_{pa(\delta)}\}
\end{equation}
and in this parameterization we have that 
\begin{equation}
\label{eq:discrete_conditional}
 p(i_\delta | \mathbf{i}_{pa(\delta)},\Theta_\Delta) = \theta_{i_\delta|\mathbf{i}_{pa(\delta)}}\ \ \  \forall \delta \in \Delta; \ \ \ \forall i_\delta \in {\mathcal I}_\delta;\ \ \   \forall \mathbf{i}_{pa(\delta)} \in {\mathcal I}_{pa(\delta)}
\end{equation}

\subsection{Distribution over continuous variables}
The  conditional distribution $p(y_\gamma|\mathbf{y}_{pc(\gamma)},\mathbf{i}_{pd(\gamma)})$ for each continuous variable index $\gamma \in \Gamma$, follows a Gaussian linear regression model\footnote{The model is reviewed in \ref{sec:BayesianGaussianLinearRegression}.} with parameters $\boldsymbol{\beta}_{\gamma|\mathbf{i}_{pd(\gamma)}}, \sigma^2_{\gamma|\mathbf{i}_{pd(\gamma)}}$, where $\boldsymbol{\beta}_{\gamma|\mathbf{i}_{pd(\gamma)}}\in \mathbb{R}^{|pc(\gamma)|+1}$ is the vector of regression coefficients (one for each continuous parent of $\gamma$ plus one for the intercept) and $\sigma^2_{\gamma|\mathbf{i}_{pd(\gamma)}}$ is the conditional variance. That is,
\begin{equation}
\label{eq:continuous_conditional}
p(y_\gamma|\mathbf{y}_{pc(\gamma)},\boldsymbol{\beta}_{\gamma|\mathbf{i}_{pd(\gamma)}}, \sigma^2_{\gamma|\mathbf{i}_{pd(\gamma)}}) = {\mathcal N}(y_\gamma|(\boldsymbol{\beta}_{\gamma|\mathbf{i}_{pd(\gamma)}})^T \mathbf{z},\sigma_{\gamma|\mathbf{i}_{pd(\gamma)}}^2)
\end{equation}
where  $\mathbf{z}=\left[\begin{array}{c}1\\\mathbf{y}_{pc(\gamma)}\end{array}\right].$

The set $\Theta_\Gamma$ includes the parameters for the model of each continuous variable:
\begin{equation}
 \Theta_\Gamma=\{(\boldsymbol{\beta}_{\gamma|\mathbf{i}_{pd(\gamma)}}, \sigma^2_{\gamma|\mathbf{i}_{pd(\gamma)}}) | \ \  \gamma \in \Gamma; \ \  \mathbf{i}_{pd(\gamma)} \in {\mathcal I}_{pd(\gamma)}\}.
\end{equation}

%It can easily be shown by induction \cite{Shachter1989,Bottcher2004} that when the local probability distributions are given as defined in \ref{eq:discrete_conditional} and \ref{eq:continuous_conditional}, the joint probability distribution for $X$ follows a CG distribution. 

Summarizing, a CGN model is defined by: (i) its structure $D$, (ii) the parameters for the discrete variables $\Theta_\Delta$, and (iii) the parameters for the continuous variables $\Theta_\Gamma.$ 

\section{\label{sec:ML}Parameter learning in conditional Gaussian networks: maximum likelihood}
%\subsection{Properties of the CG distribution}
%Next we provide some basic results which are necessary to 
%\begin{proposition}
%Let $X=(I,Y)$ follow a CG distribution and $B\subset \Gamma$, then the 
%\end{proposition}

In this section we succinctly review the results in \cite{Lauritzen1996}, providing an answer to the following question:

\begin{quote}
 If we assume that our data follows a CGN model with structure $D$, when and how can we find maximum likelihood estimates for its parameters from a sample of data $S$?
\end{quote}
 
We want to estimate the CGN parameters from a data sample $S$ that contains $n$ observations $S=\{\mathbf{x}^1,\ldots,\mathbf{x}^n\}$. Each observation $\mathbf{x}^j$ contains discrete and continuous variables, $\mathbf{x}^j=(\mathbf{i}^j,\mathbf{y}^j)$.

%Following \cite{Lauritzen1996}, given  
We introduce the following notation, where $A$ is a set of discrete variable indexes $A\subset \Delta$ and $B$ is a set of continuous variable indexes $B\subset\Gamma$:
\begin{equation*}
\begin{array}{lclp{7.5cm}}
d(\mathbf{i}_A) & = & \{j| \mathbf{i}^j_A = \mathbf{i}_A\} & Indexes of the observations in cell $\mathbf{i}_A.$ \\
n(\mathbf{i}_A) & = & |d(\mathbf{i}_A)| &  Number of elements in cell  $\mathbf{i}_A.$\\
S_{B|\mathbf{i}_A} & = & 
\left[
\begin{array}{c}
\vdots \\ (\mathbf{y}_B^j)^\top \\ \vdots
\end{array}
\right] &Matrix that contains a row $(\mathbf{y}^j_B)^\top$ for each $j\in  d(\mathbf{i}_A).$\\
s_{B|\mathbf{i}_A} & = & \sum\limits_{j\in d(\mathbf{i}_A)} \mathbf{y}_B^j & Sum of the  $\mathbf{y}_B$-values in cell $\mathbf{i}_A.$\\
\mathbf{\bar{y}}_{B|\mathbf{i}_A} & = & \frac{s_{B|\mathbf{i}_A}}{n(\mathbf{i}_A)} & Average of the $\mathbf{y}_B$-values in cell $\mathbf{i}_A.$\\
ss_{B|\mathbf{i}_A} &  = & \sum\limits_{j\in d(\mathbf{i}_A)} \mathbf{y}_B^j(\mathbf{y}_B^j)^\top & Sum of squares of the $\mathbf{y}_B$-values in cell $\mathbf{i}_A.$\\
ssd_{B|\mathbf{i}_A} & = & ss_{B|\mathbf{i}_A} - \frac{s_{B|i_A}(s_{B|\mathbf{i}_A})^\top}{n(\mathbf{i}_A)} & Sum of squares matrix of the deviations from the mean of $\mathbf{y}_B$ in cell $\mathbf{i}_A.$\\
\hat\Sigma_{B|\mathbf{i}_A} & = & ssd_{B|\mathbf{i}_A}/n(\mathbf{i}_A) & ML estimate of the covariance matrix of $\mathbf{y}_B$ in cell $\mathbf{i}_A.$\\
\end{array}
\end{equation*}

It is known (as a consequence of Proposition 6.33 in  \cite{Lauritzen1996}) that the ML parameters for this model can be assessed independently for the conditional distribution of each variable. Similarly to what is described in \cite{Perez2010}, for each variable we apply a composition of the results from \cite{Lauritzen1996} (proposition 6.9, together with the transformation formulas in page 165) to assess the ML parameters for that variable. However, in order to assess the parameters by ML, we need our sample to satisfy certain conditions. These conditions are detailed into the following definition: 

\begin{definition}[Acceptable sample]
Let $D$ be a CGN structure and $S$ be a sample. We say that $S$ is acceptable for $D$ if and only if 
\begin{enumerate}
 \item  For each discrete variable $\delta \in \Delta$, for each cell $\mathbf{i}_{ pa(\delta)\cup \{\delta\}} \in \mathcal{I}_{ pa(\delta)\cup \{\delta\}} $ we have that $n(\mathbf{i}_{ pa(\delta)\cup \{\delta\}})>0.$
 \item  For each continuous variable index $\gamma \in \Gamma$, for each cell $\mathbf{i}_{pd(\gamma)} \in \mathcal{I}_{ pd(\gamma)} $ we have that
 \begin{itemize}
  \item $n(\mathbf{i}_{pd(\gamma)})>0,$ and
  \item $ssd_{pc(\gamma)\cup\{\gamma\}|\mathbf{i}_{pd(\gamma)}}$ is positive definite.
 \end{itemize}
 %This is event is almost surely equal to the event that $n(\mathbf{i}_{pd(\gamma)}) > |pc(\gamma)|.$
\end{enumerate}
\end{definition}

Intuitively, a sample is acceptable provided that it has \emph{enough} observations of each \emph{used} cell. Thus, the larger the number of dependencies in structure $D$, the larger the size required for a sample to be acceptable. 

The following result summarizes how to assess the ML parameters of the conditional distribution of each variable, if we are given an acceptable sample.
\begin{proposition}
\label{prop:MLparams}
Provided $S$ is an acceptable sample for $D$, the following procedure assesses the parameters $\Theta_\Delta$ and $\Theta_\Gamma$ that maximize likelihood:
\begin{enumerate}
 \item For each discrete variable $\delta \in \Delta:$
 \begin{equation}
  \theta_{i_\delta | \mathbf{i}_{pa(\delta)}} = \frac{n(\mathbf{i}_{ pa(\delta)\cup \{\delta\}})}{n(\mathbf{i}_{pa(\delta)})} \ \ \ \  \ \ \forall i_\delta \in \mathcal{I}_\delta;  \ \ \ \forall \mathbf{i}_{pa(\delta)} \in \mathcal{I}_{pa(\delta)}.
 \end{equation}
 \item For each continuous variable index $\gamma \in \Gamma$, and for each $\mathbf{i}_{pd(\gamma)} \in {\mathcal I}_{pd(\gamma)}:$  
 \begin{equation}
    \boldsymbol{\beta}_{\gamma|\mathbf{i}_{pd(\gamma)}} = \begin{bmatrix}
                                              \bar{y}_{\gamma|\mathbf{i}_{pd(\gamma)}} - \mathbf{r} \cdot \mathbf{\bar{y}}_{pc(\gamma)|\mathbf{i}_{pd(\gamma)}}\\
                                              \mathbf{r}^\top
                                             \end{bmatrix}
 \end{equation}
 \begin{equation}
   \sigma^2_{\gamma|\mathbf{i}_{pa(\gamma)}} = 
   M_{\gamma,\gamma} - 
   \mathbf{r} \cdot M_{pc(\gamma),\gamma}
 \end{equation} 
 where $ M $ is the matrix $ \hat\Sigma_{pc(\gamma)\cup\{\gamma\}|\mathbf{i}_{pd(\gamma)}} $ and 
 $\mathbf{r} = M_{\gamma,pc(\gamma)} \left(M_{pc(\gamma),pc(\gamma)}\right)^{-1}$.

\end{enumerate}
\end{proposition}

\subsection{Conditional Gaussian network classifiers}

For the problem of classification, given a CGN structure $D$ and an acceptable sample $S$ for $D$, the ML parameters $\Theta_V=(\Theta_\Delta, \Theta_\Gamma)$  can be found using Proposition \ref{prop:MLparams}, completing a CGN model that can be used to classify by assessing for each possible class $x_c$, its probability given the value of all the other attributes, $p(x_c|\mathbf{x}_{-c},\Theta_V)$, as 
\begin{equation}
p(x_c|\mathbf{x}_{-c},\Theta_V) = \frac{p(x_c,\mathbf{x}_{-c}|\Theta_V)}{\sum_{x'_c\in \mathcal{I}_c}p(x'_c,\mathbf{x}_{-c}|\Theta_V)}
\end{equation}
where $p(x_c,\mathbf{x}_{-c}|\Theta_V)$ can be assessed using Equations \ref{eq:CGNFactorization}, \ref{eq:discrete_conditional}, and \ref{eq:continuous_conditional} and the values of the ML parameters in $\Theta_V$.

\section{Assessing exact BA probabilities in CGNs}
\label{sec:BMA}
An alternative to estimating the parameters by ML is performing Bayesian learning over them. Bayesian learning assumes a prior probability distribution $p(\Theta_V|\xi)$ over the parameters and refines this knowledge from a data sample $S$, to obtain a posterior probability distribution $p(\Theta_V|S,\xi)$. 
\begin{equation}\label{eq:BMAppd}
p(\Theta_V|S,\xi) \propto p(\Theta_V|\xi) p(S|\Theta_V) = p(\Theta_V|\xi)\prod_{\mathbf{x}^j\in S} p(\mathbf{x}^j|\Theta_V) 
\end{equation}
After that, the classifier assesses $p(x_c|\mathbf{x}_{-c},S,\xi)$ as
\begin{equation}
\label{eq:BMAprediction1}
p(x_c|\mathbf{x}_{-c},S,\xi) \propto p(x_c,\mathbf{x}_{-c}|S,\xi)
\end{equation}
\begin{equation}
\label{eq:BMAprediction2}
p(\mathbf{x}|S,\xi) = \int_{\Theta_V}p(\mathbf{x}|\Theta_V)p(\Theta_V|S,\xi) d\Theta_V.
\end{equation}
In order to use Bayesian learning, we need that both the assessment of the posterior $p(\Theta_V|S,\xi)$ given at Equation \ref{eq:BMAppd} and prediction (Equation \ref{eq:BMAprediction2}) can be done efficiently. This can be accomplished if we define a family of probability distributions over the parameters that is conjugate to our model. 

For conditional Gaussian networks, Bøttcher proposed such a family in \cite{Bottcher2004}. In the remaining of this section we provide the details. We start by defining a distribution over the parameters, the Directed Hyper Dirichlet Normal Inverse Gamma ($\mathcal{DHDNIG}$). Then, we show that the hyperparameters of the $\mathcal{DHDNIG}$ can be efficiently updated after observing a data sample and finally we show that it is easy to assess the posterior predictive probabilities when the parameters follow a $\mathcal{DHDNIG}$. 

\subsection{The Directed Hyper Dirichlet Normal Inverse Gamma distribution}
The $\mathcal{DHDNIG}$ (detailed in definition \ref{def:DHDNIW}) assumes that the parameters of the conditional distribution of each variable in the CGN are independent. Furthermore, it assumes that  discrete variables follow a Dirichlet distribution for each configuration of its discrete parents and that continuous variables follow a normal inverse Gamma ($\mathcal{NIG}$) for each configuration of its discrete parents.

\begin{definition}[$\mathcal{DHDNIG}$]
\label{def:DHDNIW}
The parameters $\Theta_\Delta$ and $\Theta_\Gamma$ of a CGN with structure $D$ follow a $\mathcal{DHDNIG}$  distribution with (hyper)parameters $\Psi$, noted as $\mathcal{DHDNIG}(\Theta|D,\Psi)$, where
\begin{equation}
\begin{array}{rcl}
 \Psi&=&(\Psi_\Delta,\Psi_\Gamma)\\
 \Psi_\Delta& =& \{\boldsymbol{\psi}_{\delta|\mathbf{i}_{pa(\delta)}} | \ \ \forall \delta \in \Delta ; \ \ \ \forall \mathbf{i}_{pa(\delta)} \in {\mathcal I}_{pa(\delta)}\}\\
 \boldsymbol{\psi}_{\delta|\mathbf{i}_{pa(\delta)}} &=& \{\psi_{i_\delta|\mathbf{i}_{pa(\delta)}}>0 | \ \ \ \forall i_\delta \in \mathcal{I}_\delta\}\\
 \Psi_\Gamma &=& \{\boldsymbol{\psi}_{\gamma|\mathbf{i}_{pd(\gamma)}} | \ \ \forall \gamma \in \Gamma ; \ \ \ \forall \mathbf{i}_{pd(\gamma)} \in {\mathcal I}_{pd(\gamma)}\}\\
 \boldsymbol{\psi}_{\gamma|\mathbf{i}_{pd(\gamma)}}&=& (\boldsymbol{\mu}_{\gamma|\mathbf{i}_{pd(\gamma)}},V_{\gamma|\mathbf{i}_{pd(\gamma)}},\rho_{\gamma|\mathbf{i}_{pd(\gamma)}},\phi_{\gamma|\mathbf{i}_{pd(\gamma)}})
\end{array} 
\end{equation}
provided that
\begin{itemize}
 \item For each discrete variable index $\delta \in \Delta$ and for each cell of its parents ($\mathbf{i}_{pa(\delta)} \in {\mathcal I}_{pa(\delta)}$), the parameters of the multinomial $\boldsymbol{\theta}_{\delta|\mathbf{i}_{pa(\delta)}}$ follow a Dirichlet distribution with parameters $\boldsymbol{\psi}_{\delta|\mathbf{i}_{pa(\delta)}}$:
\begin{equation}
p(\boldsymbol{\theta}_{\delta|\mathbf{i}_{pa(\delta)}}|\Psi) = \mathcal{D}(\boldsymbol{\theta}_{\delta|\mathbf{i}_{pa(\delta)}}|\boldsymbol{\psi}_{\delta|\mathbf{i}_{pa(\delta)}})
\end{equation}
\item For each continuous variable index $\gamma \in \Gamma$ and for each cell of its discrete parents ($\mathbf{i}_{pd(\gamma)} \in {\mathcal I}_{pd(\gamma)}$), the parameters of the Gaussian linear regression $\boldsymbol{\beta}_{\gamma|\mathbf{i}_{pd(\gamma)}}, \sigma^2_{\gamma|\mathbf{i}_{pd(\gamma)}}$ follow a $\mathcal{NIG}$ with hyperparameters $\boldsymbol{\psi}_{\gamma|\mathbf{i}_{pd(\gamma)}}= (\boldsymbol{\mu}_{\gamma|\mathbf{i}_{pd(\gamma)}},V_{\gamma|\mathbf{i}_{pd(\gamma)}},\rho_{\gamma|\mathbf{i}_{pd(\gamma)}},\phi_{\gamma|\mathbf{i}_{pd(\gamma)}})$:
\begin{multline}
 p(\boldsymbol{\beta}_{\gamma|\mathbf{i}_{pd(\gamma)}}, \sigma^2_{\gamma|\mathbf{i}_{pd(\gamma)}}|\Psi) =\\ = \mathcal{NIG}(\boldsymbol{\beta}_{\gamma|\mathbf{i}_{pd(\gamma)}}, \sigma^2_{\gamma|\mathbf{i}_{pd(\gamma)}}|\boldsymbol{\mu}_{\gamma|\mathbf{i}_{pd(\gamma)}},V_{\gamma|\mathbf{i}_{pd(\gamma)}},\rho_{\gamma|\mathbf{i}_{pd(\gamma)}},\phi_{\gamma|\mathbf{i}_{pd(\gamma)}})
\end{multline}
\end{itemize}

\end{definition}

\subsection{Learning}
Proposition~\ref{prop:Learning} summarizes how the hyperparameters of a $\mathcal{DHDNIG}$  distribution should be updated provided that we observe a sample $S$. 
\begin{proposition}
\label{prop:Learning} 
Given a CGN structure $D$ and assuming the parameters follow a $\mathcal{DHDNIG}(\Theta|D,\Psi)$, the posterior probability over parameters $p(\Theta|D,S,\Psi)$ follows a $\mathcal{DHDNIG}(\Theta|D,\Psi')$, where for each $\delta \in \Delta$ and , each $i_{\delta} \in \mathcal{I}_{\delta}$ and each $\mathbf{i}_{pa(\delta)} \in \mathcal{I}_{pa(\delta)}$ we have 
\begin{eqnarray}
\psi'_{i_\delta|\mathbf{i}_{pa(\delta)}} = \psi_{i_\delta|\mathbf{i}_{pa(\delta)}} + n(\mathbf{i}_{pa(\delta)\cup\{\delta\}}) 
\end{eqnarray}
and for each $\gamma \in \Gamma$  and each $\mathbf{i}_{pd(\gamma)} \in \mathcal{I}_{pd(\gamma)}$ we have 
\begin{small}
\begin{alignat}{4}
&V'_{\gamma|\mathbf{i}_{pd(\gamma)}} & = &\  [V_{\gamma|\mathbf{i}_{pd(\gamma)}}^{-1} +  ss_{pc(\gamma)\cup\{\gamma\}|\mathbf{i}_{pd(\gamma)}}]^{-1},\\
&\boldsymbol{\mu}'_{\gamma|\mathbf{i}_{pd(\gamma)}} \ & = & \ V_{\gamma|\mathbf{i}_{pd(\gamma)}}'(V_{\gamma|\mathbf{i}_{pd(\gamma)}}^{-1}\boldsymbol{\mu}_{\gamma|\mathbf{i}_{pd(\gamma)}} + S_{pc(\gamma)\cup\{\gamma\}|\mathbf{i}_{pd(\gamma)}}^T S_{\gamma|\mathbf{i}_{pd(\gamma)}}),\\
&\rho'_{\gamma|\mathbf{i}_{pa(\gamma)}} &= &\ \rho_{\gamma|\mathbf{i}_{pa(\gamma)}} + \frac{n(\mathbf{i}_{pd(\gamma)})}{2},\\
&\phi'_{\gamma|\mathbf{i}_{pa(\gamma)}} & = & \  \phi_{\gamma|\mathbf{i}_{pa(\gamma)}} + \frac{1}{2}\big[
\boldsymbol{\mu}^\top_{\gamma|\mathbf{i}_{pd(\gamma)}} V_{\gamma|\mathbf{i}_{pd(\gamma)}}^{-1}\boldsymbol{\mu}_{\gamma|\mathbf{i}_{pd(\gamma)}} 
+ ss_{\gamma|\mathbf{i}_{pa(\gamma)}} \\
&&& \hspace{2cm}- {\boldsymbol{\mu}'}_{\gamma|\mathbf{i}_{pd(\gamma)}}^\top {V'}_{\gamma|\mathbf{i}_{pd(\gamma)}}^{-1} \boldsymbol{\mu'}_{\gamma|\mathbf{i}_{pd(\gamma)}} \big] \nonumber.
\end{alignat}
\end{small}
\end{proposition}
The result follows from the fact that the $\mathcal{DHDNIG}$ factorizes over the structure and from the results for multinomial distributions and Gaussian linear regressions provided in \ref{sec:BayesianMultinomial} and \ref{sec:BayesianGaussianLinearRegression}. 
\subsection{Predicting}
Proposition~\ref{prop:Predicting} shows how we can determine the probability of a new observation in a CGN whose parameters follow a $\mathcal{DHDNIG}$  distribution.
\begin{proposition}
\label{prop:Predicting}
Given a CGN structure $D$ and assuming the parameters follow a $\mathcal{DHDNIG}(\Theta|D,\Psi)$, the probability of an observation $\mathbf{x}$ can be assessed as 
\begin{equation}
p(\mathbf{x}|D,\Psi)=\prod_{\delta\in \Delta}p(i_\delta|\mathbf{i}_{pa(\delta)},D,\Psi)   \prod_{\gamma\in \Gamma}p(y_\gamma|\mathbf{y}_{pc(\delta)},\mathbf{i}_{pc(\delta)} ,D,\Psi)                                                                                                                                    \end{equation}
where 
\begin{equation}
p(i_\delta|\mathbf{i}_{pa(\delta)},D,\Psi) = 
\frac{\psi_{i_\delta|\mathbf{i}_{pa(\delta)}}}
{\sum_{i'_\delta\in \mathcal {I}_\delta} \psi_{i'_\delta|\mathbf{i}_{pa(\delta)}}},
\end{equation} and 
\begin{multline}
p(y_\gamma|\mathbf{y}_{pc(\gamma)},\mathbf{i}_{pd(\gamma)} ,D,\Psi) = \\
= St\left(y_\gamma | 2\rho_{\gamma|\mathbf{i}_{pa(\gamma)}},
\mathbf{z}^\top \boldsymbol{\mu}_{\gamma|\mathbf{i}_{pa(\gamma)}},
\frac{\phi_{\gamma|\mathbf{i}_{pa(\gamma)}}}{\rho_{\gamma|\mathbf{i}_{pa(\gamma)}}}
  (1+\mathbf{z}^\top V_{\gamma|\mathbf{i}_{pa(\gamma)}} \mathbf{z})\right)
\end{multline}
where $\mathbf{z}=\left[\begin{array}{c}1\\\mathbf{y}_{pc(\gamma)}^\top\end{array}\right].$
\end{proposition}

Again, the result follows from the fact that the $\mathcal{DHDNIG}$ factorizes over the structure and from the results for multinomial distributions and Gaussian linear regressions provided in \ref{sec:BayesianMultinomial} and \ref{sec:BayesianGaussianLinearRegression}. 
%\subsubsection{Uninformative hyperparameter setting}

\subsection{Suggested hyperparameters}\label{sec:hyper}
Along the line proposed in \cite{Bottcher2004}, we propose to use the following prior, inspired in assuming that all the variables are independent. 
\begin{enumerate}
\item For each discrete variable, initialize the Dirichlet distribution hyperparameters to a small positive value. For our experiments we have chosen $0.01:$%s fLet star for the parameter of the discrete variable we asses the parameters as follows,
\begin{equation}
\psi_{i_\delta|\mathbf{i}_{pa(\delta)}} = 0.01 \ \ \ \ \ \ \ \ \ \ \ \forall \delta \in \Delta\ ;\  \forall i_\delta \in \mathcal{I}_\delta \ ;\ \forall \mathbf{i}_{pa(\delta)} \in \mathcal{I}_{pa(\delta)}
\end{equation}
%experiments we will use the prior proposed by Bøttcher in In which we asses the parameters for Normal Inverse Wishart in the case of the normal distribution and then use the %learned prior to asses the $\mathcal{DHDNIG}$ prior. 
%it is done in order to not give to much weight to the prior, but assessing our believe in that every possible value of the variable is  uniformly distributed.

%In order to asses the parameters of the prior distribution we suppose that the continuous part of the distribution follows the following prior.
\item For each continuous variable, initialize the $\mathcal{NIG}$ hyperparameters to 
\begin{alignat}{4}
&V_{\gamma|\mathbf{i}_{pd(\gamma)}} & = &  \left( \begin{matrix}
1 + \bar{\mathbf{ y}}_{pc(\gamma)|\mathbf{i}_{pd(\gamma)}}^\top K^{-1}_{pc(\gamma)}\bar{\mathbf{y}}_{pc(\gamma)|\mathbf{i}_{pd(\gamma)}} & -\bar{\mathbf{y}}_{pc(\gamma)|\mathbf{i}_{pd(\gamma)}}^\top K^{-1}_{pc(\gamma)} \\
-K^{-1}_{pc(\gamma)}\bar{\mathbf{y}}_{pc(\gamma)|\mathbf{i}_{pd(\gamma)}} & K^{-1}_{pc(\gamma)}
\end{matrix}\right),\\
&\boldsymbol{\mu}_{\gamma|\mathbf{i}_{pd(\gamma)}}  & = & (\bar{y}_{\gamma},0,\ldots ,0),\\
&\rho_{\gamma|\mathbf{i}_{pd(\gamma)}} &= &1.1 + \frac{|pc(\gamma)|}{2},\\
&\phi_{\gamma|\mathbf{i}_{pd(\gamma)}} & = & \frac{\hat{\Sigma}_\gamma}{2}.
\end{alignat}
where $\bar{\textbf{y}}$ is the empirical mean, and $K_{pa(\gamma)}$ is a diagonal matrix containing the value $\hat\Sigma_{\gamma'}$ (the variance of variable $\gamma'$) in its diagonal for each variable $\gamma' \in pa(\gamma).$ 
\end{enumerate}

\subsection{Algorithm and discussion}

Given a sample $S$ and a structure $D$ the procedure to create a classification model using BA starts by initializing the hyperparameters of the prior $\mathcal{DHDNIG}$ distribution $\Psi$ as suggested in section~\ref{sec:hyper}. After that, it uses Proposition \ref{prop:Learning} to assess the posterior distribution $\Psi'$. Finally, it asseses the probability of each class $x_c$ given the posterior distribution, $p(x_c|\mathbf{x}_{-c}, \Psi'),$ as
\begin{equation}
p(x_c|\mathbf{x}_{-c},\Psi') = \frac{p(x_c,\mathbf{x}_{-c}|\Psi)}{\sum_{x'_c\in \mathcal{I}_c}p(x'_c,\mathbf{x}_{-c}|\Psi')}
\end{equation}
where $p(x_c,\mathbf{x}_{-c}|\Psi')$ is calculated using Proposition \ref{prop:Predicting}.

%\textcolor{red}{Describe how the previous results in the section can be used to create a classifier (which is fairly obvious).}

%
%\textcolor{red}{Discuss how the results in this section generalize those in the PGM paper and connect with the literature.}

%\textcolor{red}{We should talk a bit about the code here also. The source code for the experiments is available at \cite{Bellon2012b}.
%%\emph{http://www.iiia.csic.es/~cerquide/pypermarkov}

%\textcolor{red}{Describe how the previous results in the section can be used to create a classifier (which is fairly obvious).}
In  \cite{Bellon2012}, we presented the Hyper Dirichlet Normal Inverse Wishart distribution $\mathcal{HDNIW}$ as a tool to perform exact Bayesian averaging over parameters in Markov networks when the structure was decomposable. The classifiers presented in this section are a generalization of those presented in \cite{Bellon2012}, since (i) a decomposable Markov network can be represented as a Bayesian network, and (ii) the $\mathcal{HDNIW}$ can be reparameterized as a $\mathcal{DHDNIG}$ in that Bayesian network. For the same reason, the results in \cite{Corander2012} can be seen as a particular case of the results presented in this section.

%\textcolor{red}{Discuss how the results in this section generalize those in the PGM paper and connect with the literature.}

%The source code for the experimentation is avaible at \cite{Bellon2012b}. The learning process is done by an inducer object instance which has two main components, an structure learner and a parameter learner.  First of all a structure is learned by the inducer and then a distribution is learned for each variable of the structure. The returning result is a model with the capability of classify instances. 

%There are two main different distribution types. Those distributions that can be used as priors inherit from \emph{conjugate\_cond\_distribution} class, and different prior parameters can be chosen for this distributions, the most important methods of this distributions are \emph{incorporate\_evidence}, which calculates the posterior distribution from some given data, and \emph{get\_predictive\_posterior}, that returns a predictive posterior distribution. In the other hand predictive distribution inherit their properties from the \emph{cond\_distribution} class. The used distributions are determined by the inducer options of the inducer and the call of their methods is done through the model function. 
%\textcolor{red}{We should talk a bit about the code here also. The source code for the experiments is available at \cite{Bellon2012b}.
%\emph{http://www.iiia.csic.es/~cerquide/pypermarkov}

\section{Experimental comparison}
\label{sec:Experimental}
In this section we compare CGN classifiers that use ML and Bayesian learning methods for the parameters. A recent thorough analysis of CGN classifiers based in ML is provided by Pérez in \cite{Perez2010}. There, several heuristic structure learning algorithms are compared, concluding that wrapper based algorithms based on Join Augmented Naïve Bayes (JAN) structures perform better than the rest. Since we are only interested in comparing parameter learning strategies, we will restrict our comparison to these structures. Furthermore, we will use the same datasets in \cite{Perez2010} for the comparison. Next, we quickly review JAN structures, and the heuristic procedures described in \cite{Perez2010} to learn them.

%in rely on follow the  of We run a set of test composed by three different structure learning algorithms proposed by  based in JAN structures. 

\subsection{Join Augmented Naïve Bayes structures}
\label{sec:JAN}
\begin{figure}
\centering
\subfigure[Naïve Bayes classifier]{\label{fig:NB}\includegraphics[width=0.45\textwidth]{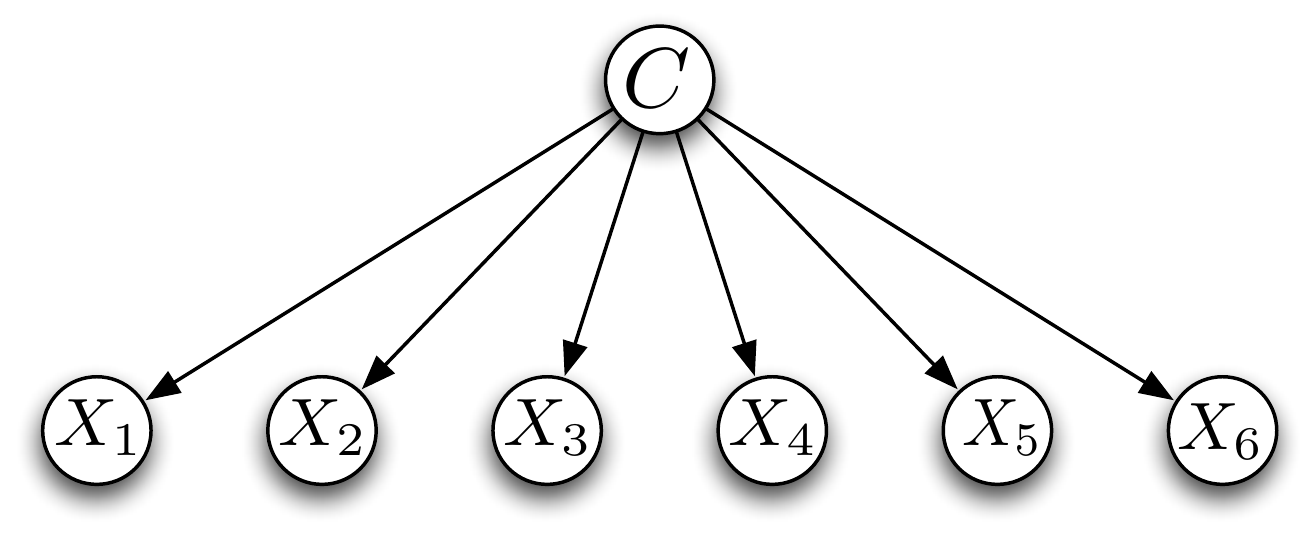}}  \ \ \ \ \ \ 
\subfigure[Join Augented Naïve Bayes classifier]{\label{fig:JAN}\includegraphics[width=0.45\textwidth]{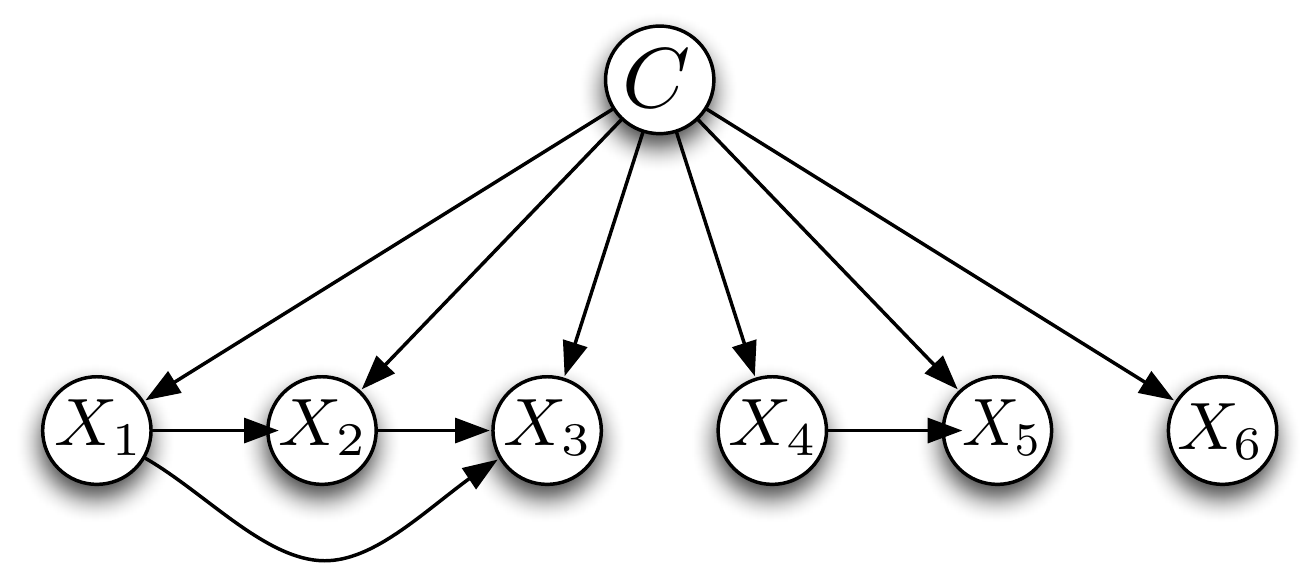}}\\
\caption{Example of a Naïve Bayes classifier and a Join Augmented Naïve Bayes classifier with six variables.}
\end{figure}

The Naïve Bayes classifier makes the assumption that each of the attributes is independent from all the other attributes given the class. The encoding of this strong independence assumption as a BN appears in Figure~\ref{fig:NB}. Unfortunately, data rarely satisfies the assumption. Thus, better BN classifiers can be obtained by introducing dependencies between the attributes. JAN structures can be seen as naive Bayes classifiers where the variables are partitioned into groups. The new assumption is that each group of variables is independent from all other groups given the class. However, no independencies are assumed inside each of the groups. An example of BN encoding the dependencies of a JAN with three groups $\{X_1,X_2,X_3\},$ $\{X_4,X_5\},$ and $\{X_6\}$ is shown in Figure~\ref{fig:JAN}. 

Wrapper algorithms \cite{Kohavi1997} have a long tradition in machine learning. For the task of structure learning, the wrapper algorithm analyzes several structures, using the training set to evaluate their performance and selecting the structure which maximizes the performance measure. In \cite{Perez2010}, the accuracy of the structure in a 10 fold cross validation over the training set in used as performance measure.

The wrapper algorithms proposed in \cite{Perez2010} follow a greedy search approach summarized in Algorithm~\ref{alg:WrapperSearch} and differ only on the initial structure and the set of candidates considered at each step of the algorithm. Three different algorithms are proposed, the forward wrapper Gaussian joint augmented naïve Bayes (fwGJAN), the backwards wrapper Gaussian joint augmented naïve Bayes (bwGJAN) and the wrapper condensed Gaussian joint augmented naïve Bayes (wcGJAN). 

The fwGJAN algorithm starts with a structure containing only the class node. At each iteration, the candidate set is constructed from the best structure so far by considering the addition of each attribute not present in the current structure, either inside one of the already existing variable groups or creating a new group of its own. 

The bwGJAN algorithm starts with a naïve Bayes structure with all the attributes in the dataset.  At each iteration, the candidate set is constructed from the best structure so far by (i) considering the removal of a single variable in the structure and (ii) joining two groups of variables in the structure.   

The wcGJAN algorithm starts with a complete structure. At each iteration, the candidate set is constructed from the best structure so far by removing a variable from the classifier. 

\begin{algorithm}
\caption{\label{alg:WrapperSearch}General wrapper structure search algorithm}
\begin{algorithmic}
\Function{LearnStructure}{$S$}
\State $BestCandidate \leftarrow $ InitialStructure()
%\State $BestValue \leftarrow $ Evaluate($BestCandidate,S$)
\Repeat
    \State $BestStructure \leftarrow BestCandidate$
    \State $Candidates \leftarrow $ GetCandidates($BestStructure$)
    \State $BestCandidate \leftarrow \underset{D\in Candidates}{\operatorname{argmax}} \text{Evaluate}(D,S)$
    %\ForAll{$D \in Candidates$}
    %    \If{Evaluate($D,S$} $> BestValue$}
    %        $Improvement \leftarrow True$
    %        
    %    \Endif
    %\EndFor
\Until{$BestStructure$ is better than $BestCandidate$ }
\State \textbf{return} $BestStructure$
\EndFunction
\end{algorithmic}
\end{algorithm}

\begin{table}
 
\end{table}

\subsection{Comparison Results}

Following \cite{Perez2010}, we use 9 UCI repository data sets, which only contain continuous predictor variables. 
%We have to take into account that most part of this repository is preprocessed \cite{Kohavi1995}: there are few irrelevant or redundant variables an little noise\cite{Putten2004}. In consequence, it is more difficult to obtain statistically significant differences between the results of the algorithms in this type of data sets \cite{Putten2004}. 
The characteristics of each dataset appear in Table~\ref{tab:Datasets}.

\begin{table}
\begin{center}
\begin{tabular}{c|c|c|c|c}
\#	&Dataset 		& \# classes 	& \# variables 	& \# observations \\\hline
1	& Balance 		& 3 				& 4 					& 625 \\\hline
2	& Block		& 5 				& 10					& 5474 \\\hline
3	& Haberman 	& 2 				& 3					& 307 \\\hline
4	& Iris			& 3 				& 4 					& 150 \\\hline
5	& Liver 		& 2 				& 6 					& 345 \\\hline
6	& Pima 		& 2 				& 8 					& 768 \\\hline
7	& Vehicle 		& 4 				& 19					& 846 \\\hline
8	& Waveform 	& 3 				& 21					& 5000 \\\hline
9	& Wine 		& 3 				& 13					& 179 \\\hline
\end{tabular}
{\caption{This table shows the characteristics of the differents datasets.}\label{tab:Datasets}}
\end{center}
\end{table}

For each dataset, we ran 10 repetitions of 10-fold cross validation and assessed the accuracy: the ratio of the number of data classified correctly to the total number of data classified; and conditional log-likelihood (CLL): the sum of the logarithm of the probability assigned by the classifier to the real class. While accuracy gives us information about how many instances are correctly classified, CLL measures how accurately the probabilities for each class are estimated, which is very relevant for adequate decision making.

For each repetition of the experiment we have used the three different JAN structure learning algorithms proposed in Section~\ref{sec:JAN}. Each proposed structure is evaluated in terms of the accuracy obtained using ML for learning the parameters of the corresponding classifier. The best structure is used for the final classifier, whose parameters are learned using both ML and BMA.  Since we are interested in classifiers that provide good results when data is scarce, we have performed the experiments two times, the first time learning from the complete training set and the second time discarding 80\% of the data in the training set.

 \begin{table}
 \begin{small}
 \centering
 \begin{tabular}{c|c|c|c|c|c|c|c|c|c|c}
Datasets \#&    1&   2&  3&  4&  5&  6&  7&  8& 9& Total\\\hline
CLL 100\% &\checkmark&\checkmark&\checkmark&\cross&\checkmark&\checkmark&\cross&\checkmark&\checkmark&7/2\\\hline
CLL 20\% &\checkmark&\checkmark&\checkmark&\checkmark&\checkmark&\checkmark&\checkmark&\checkmark&\checkmark&9/0\\\hline
ACC 100\% &\cross&\cross&\checkmark&\cross&\checkmark&\checkmark&\cross&\checkmark&\cross&4/5\\\hline
ACC 20\% &\cross&\cross&\checkmark&\checkmark&\cross&\checkmark&\cross&\checkmark&\checkmark&5/4\\\hline
\end{tabular} 
\caption{Summary of test for bwCGN structures. \checkmark\hspace{1em}denotes a winning for BA, while \cross\hspace{1em}denotes a loss and \equal\hspace{1em}a tie.}\label{tab:WBsum}
\end{small}
\end{table}

In order to analyze the results we have performed a Mann-Whitney paired test between BA and ML for each dataset and structure. We have recorded a parameter learning method as winner every time that the test was significant with a sig\-ni\-fi\-cance level of $\alpha = 5\%$ and its rank was greater than its counterpart. If the test was not significant we recorded a draw. We provide a summary of winnings and losses for each structure in Tables \ref{tab:WBsum}-\ref{tab:WFsum} .

%\begin{table}
%\begin{tabular}{c|c|c|c}
%Wins		& wbGJAN 	& wcGJAN		& wfGJAN		\\\hline
%BMA 		&4			&4			&4	 		\\\hline
%ML		&5			&5			&4			\\\hline
%
%\end{tabular}
%{\caption{Summary of datasets of wins of each method using accuracy and 100\% of learning data.}\label{tab:acc100}}
%\end{table}
%
%\begin{table}
%\begin{tabular}{c|c|c|c}
%Wins		& wbGJAN 	& wcGJAN		& wfGJAN		\\\hline
%BMA 		&7			&7			&7	 		\\\hline
%ML		&2			&2			&2			\\\hline
%
%\end{tabular}
%{\caption{Summary of datasets of wins of each method using CLL and 100\% of learning data.}\label{tab:cll100}}
%\end{table}
%
%\begin{table}
%\begin{tabular}{c|c|c|c}
%Wins		& wbGJAN 	& wcGJAN		& wfGJAN		\\\hline
%BMA 		&5			&5			&4	 		\\\hline
%ML		&4			&3			&5			\\\hline
%
%\end{tabular}
%{\caption{Summary of datasets of wins of each method using accuracy and 20\% of learning data.}\label{tab:acc20}}
%\end{table}
%
%\begin{table}
%\begin{tabular}{c|c|c|c}
%Wins		& wbGJAN 	& wcGJAN		& wfGJAN		\\\hline
%BMA 		&9			&9			&9	 		\\\hline
%ML		&0			&0			&0			\\\hline
%
%\end{tabular}
%{\caption{Summary of datasets of wins of each method using CLL and 20\% of learning data.}\label{tab:cll20}}
%\end{table}
\begin{table}
\centering
\begin{small}

\begin{tabular}{c|c|c|c|c|c|c|c|c|c|c}
Datasets \#&	1&	2&	3&	4&	5&	6&	7&	8& 9&Total\\\hline
CLL 100\% &\checkmark&\checkmark&\checkmark&\cross&\checkmark&\checkmark&\cross&\checkmark&\checkmark&7/2\\\hline
CLL 20\% &\checkmark&\checkmark&\checkmark&\checkmark&\checkmark&\checkmark&\checkmark&\checkmark&\checkmark&9/0\\\hline
ACC 100\% &\cross&\cross&\checkmark&\checkmark&\cross&\checkmark&\cross&\checkmark&\checkmark&5/4\\\hline
ACC 20\% &\cross&\cross&\checkmark&\checkmark&\checkmark&\checkmark&\cross&\checkmark&\equal&5/3\\\hline
\end{tabular} 
\caption{Summary of test for wcCGN structures. \checkmark\hspace{1em}denotes a winning for BA, while \cross\hspace{1em}denotes a loss and \equal\hspace{1em}a tie.}\label{tab:WCsum}
\end{small}
\end{table}
\begin{table}
\begin{small}
\centering
\begin{tabular}{c|c|c|c|c|c|c|c|c|c|c}
Datasets \#&	0&	1&	2&	3&	4&	5&	6&	7&	8& Total\\\hline
CLL 100\% &\cross&\checkmark&\checkmark&\cross&\checkmark&\checkmark&\checkmark&\checkmark&\checkmark&7/2\\\hline
CLL 20\% &\checkmark&\checkmark&\checkmark&\checkmark&\checkmark&\checkmark&\checkmark&\checkmark&\checkmark&9/0\\\hline
ACC 100\% &\cross&\cross&\equal&\cross&\checkmark&\checkmark&\cross&\checkmark&\checkmark&4/4\\\hline
ACC 20\% &\cross&\cross&\checkmark&\checkmark&\cross&\cross&\cross&\checkmark&\checkmark&4/5\\\hline
\end{tabular} 
\caption{Summary of test for fwCGN structures. \checkmark\hspace{1em}denotes a winning for BA, while \cross\hspace{1em}denotes a loss and \equal\hspace{1em}a tie.}\label{tab:WFsum}
\end{small}
\end{table}  

%Our experiments confirm the results of \cite{AritzPerez2010}, showing that wcGJAN ranks better than the other two alternatives for learningstructures.
The results are similar for the three structure learning methods. In most of the datasets, the classifiers learned using BA provide a higher CLL than those learned using ML. That is, BA provides more accurate probability predictions. The results for accuracy seem very similar for BA and ML. Furthermore, as shown by previous research \cite{Cerquides2005}, the advantages of using BA are clearer as we reduce the amount of learning data. In the next section we will see that this is confirmed in a problem with highly scarce data.

%In summary, we can see that Bayesian Model Averaging improves the results obtained by classifiers learned using ML methods. 

\section{CGN classifiers for early diagnosis of ovarian cancer from mass spectra}
\label{sec:Cancer}
%As noted in \cite{Tur2011}, whenever the number of variables in our dataset is large with respect to the number of observations that we have, the , 
In this section we compare CGN classifiers using ML and BA for the task of early prediction of ovarian cancer from mass spectra. Mass spectrometry is a scientific technique for measuring the mass of ions. For clinical purposes, the mass spectrum of a sample of blood or other substance of the patient can be obtained. Mass spectra provide a wealth of information about the molecules present in the sample. In particular, each mass spectrum can be understood as a huge histogram, where the number of molecules observed in the sample is reported for each mass/charge quotient (m/z). The objective pursued is to learn to automatically distinguish mass spectra of ovarian cancer patients from those of control individuals.

The data used has been obtained from the NIH and contains high resolution spectrograms coming from surface-enhanced laser desorption/ionization time of flight mass spectrometry (SELDI-TOF MS). %The objective is to distinguish spectrograms coming from cancer patients from those coming from control individuals.
%Spectrogram explanation
The dataset contains a total of 216 spectrograms, 121 from cancer patients and 95 controls. The m/z values do not coincide along the different spectrograms. Thus, to create the variables, the m/z axis data has been discretized into different bins, creating a variable for each bin, for a total of 11300 variables. Thus, the number of variables largely exceeds the number of observations. For each spectrogram, the average of the values of each bin has been assigned to that bin's variable. 

%The pancreatic cancer dataset contains a total of 181 spectrograms, 101 from cancer patients and 80 controls. Each spectrogram is defined by 6771 variables which in this case are aligned, so no discretization is needed.

\subsection{Structures}
Due to the large number of attributes, none of the algorithms for structure learning reviewed in the previous section can be used. Instead, we have used two different families of structures for the CGN. Both are based on the hypothesis that those variables that represent close m/z relations are more likely to have large correlations than those whose m/z values are further away. 
%The models are defined by the form of the global covariance matrix which will produce a structure for the graph of the model. In both the restriction of the connections is related to the ordering of the variables. As our variables corresponds mass/charge ratios, the natural ordering proposed by them is used.

\begin{figure}
\centering
\subfigure[$k$-BOX model]{\label{fig:kboxgraph}\includegraphics[width=0.45\textwidth]{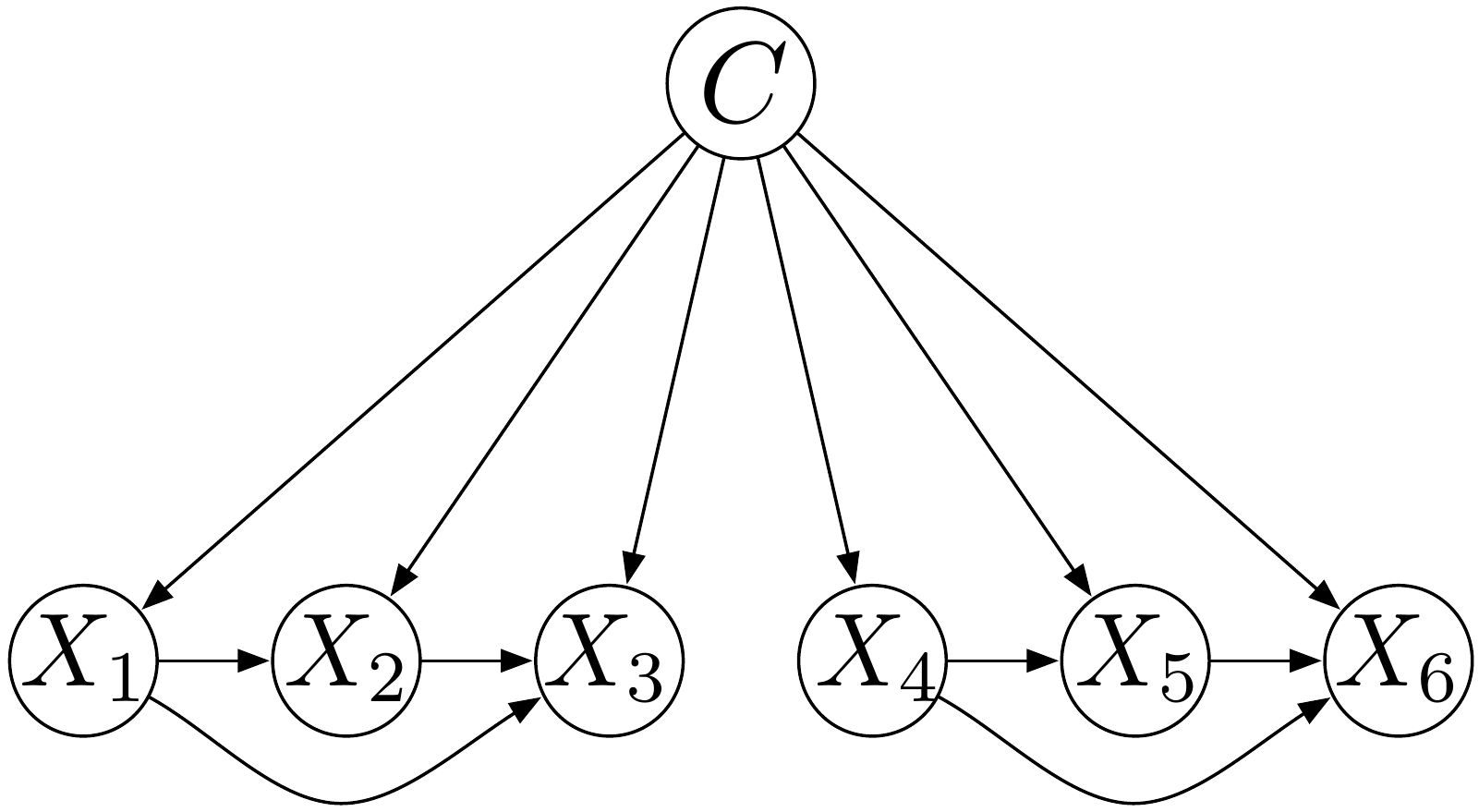}} \ \ \  \subfigure[$k$-BOX model]{\label{fig:kboxmat}\includegraphics[viewport=0 0 280 280,width=0.4\textwidth,clip]{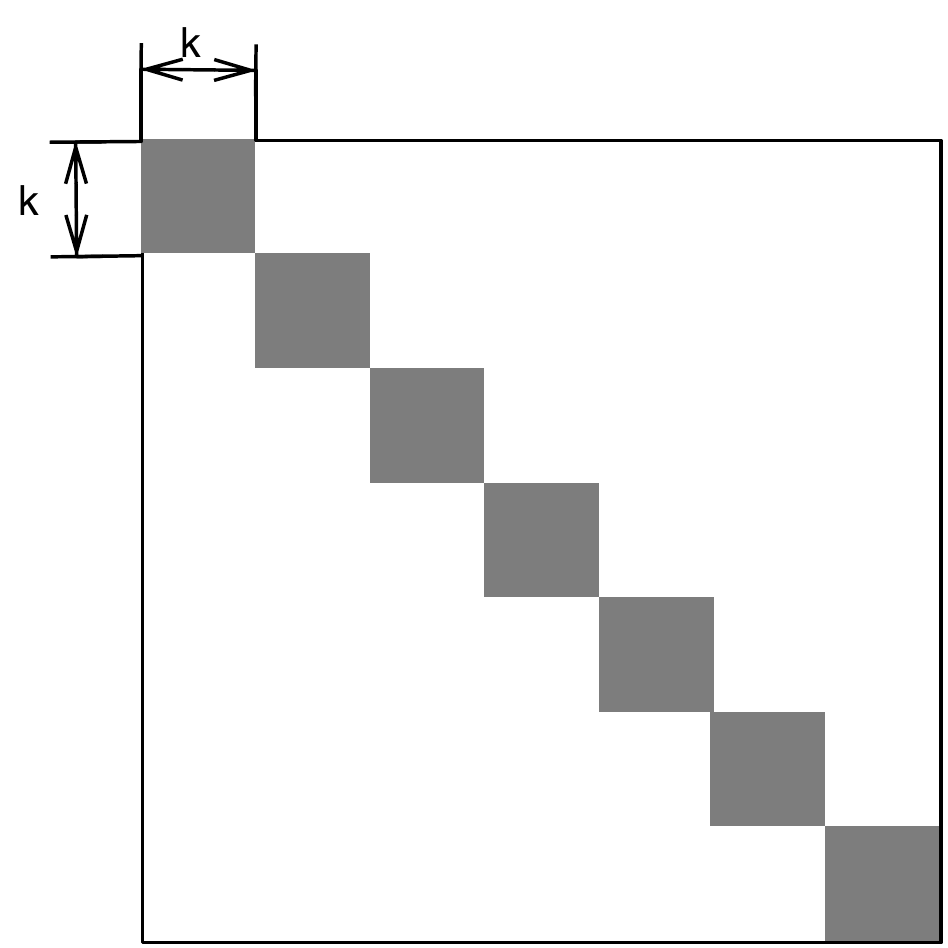}}
\caption{In a) we show the graph of a 3-BOX model for six different variables. In b) we show the connectivity matrix for a general K-BOX model.}
\end{figure}

A $k$-BOX structure can be defined over an ordered set of variables $V=\langle X_1,\ldots,X_n\rangle$. We say that a set of $k$ variables $V'$ is contiguous whenever $V'=\{X_j,\ldots,X_{j+k}\}.$ %It is a restriction of \emph{semi Naïve Bayes} \cite{Perez2006}, also known as \emph{JAN} \cite{Perez2010}. 
%In a JAN structure the variables are joined in groups to form multivariate distributions. 
The $k$-BOX structure divides the variables into disjoint contiguous sets of $k$ variables. The network structure can be seen in Figure~\ref{fig:kboxgraph} and the corresponding covariance matrix in Figure~\ref{fig:kboxmat}.
%Thus is, we generate a collection of cliques $\{C_i | i\in\mathbb{N}\}$ with $C_i=\{X_{(i-1)k+1},\ldots,C_{ik}\}$. Let's $chld(X)$ denote the set of childrens of a variable $X$, then $chld(X_i)=\{X_j |X_i \in C_k,X_j\in C_k, i<j\}$ for some clique $C_k$. 
In our case the ordering is provided by the m/z value. 

% In Figure \ref{kboxmat} we can see an scheme of the covariance matrix of the joint gaussian distribution formed by all the continuous variables given the class.  we can see an example of a k-BOX structure with $k=3$ where $X_1,X_2,X_3$ forms a clique and $X_4,X_5,X_6$ another one. 

\begin{figure}
\centering
\subfigure[$k$-BAND model]{\label{fig:kbandgraph}\includegraphics[width=0.45\textwidth]{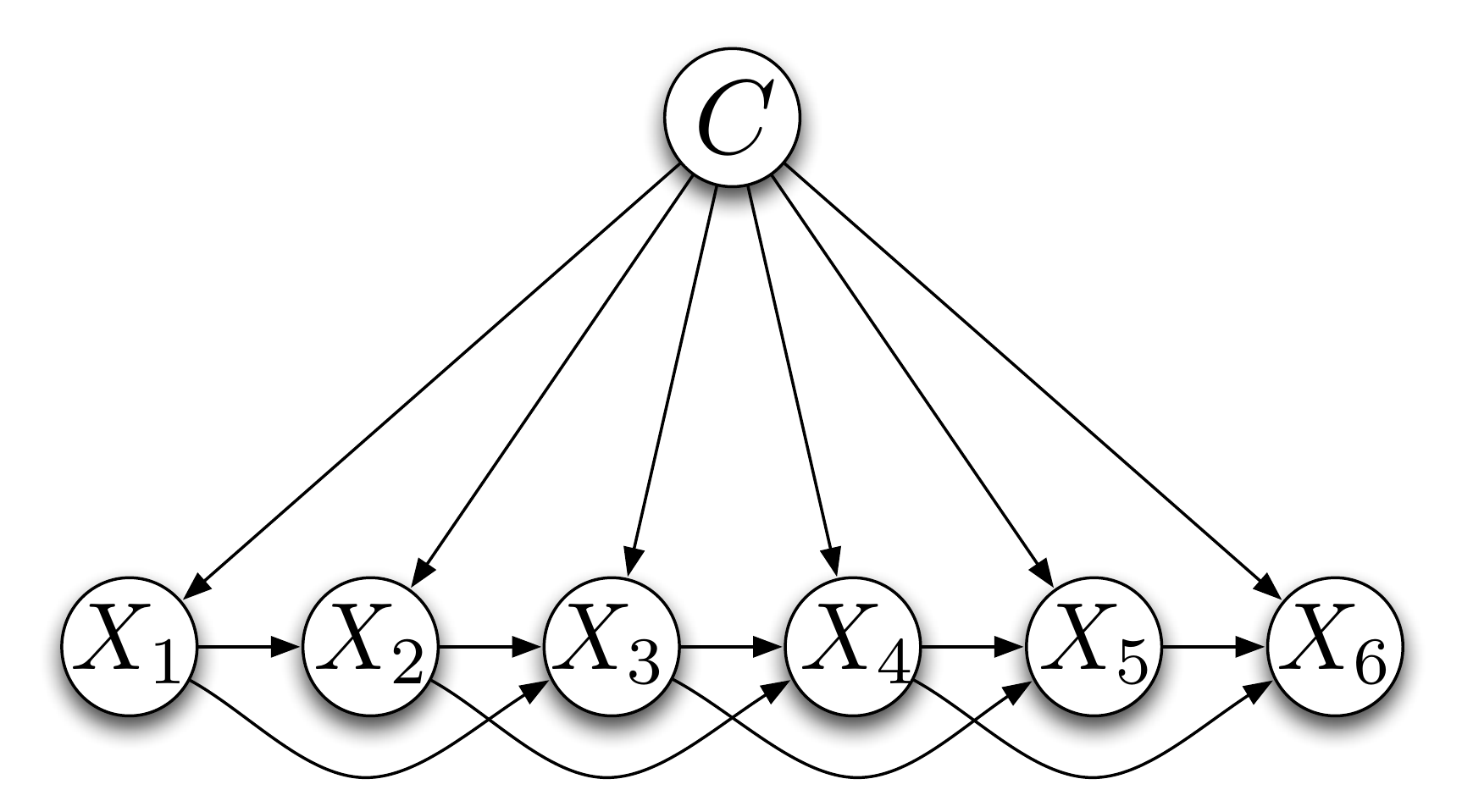}}
\ \ \ 
\subfigure[$k$-BAND model]{\label{fig:kbandmat}\includegraphics[viewport=3 1 270 270,width=0.4\textwidth,clip]{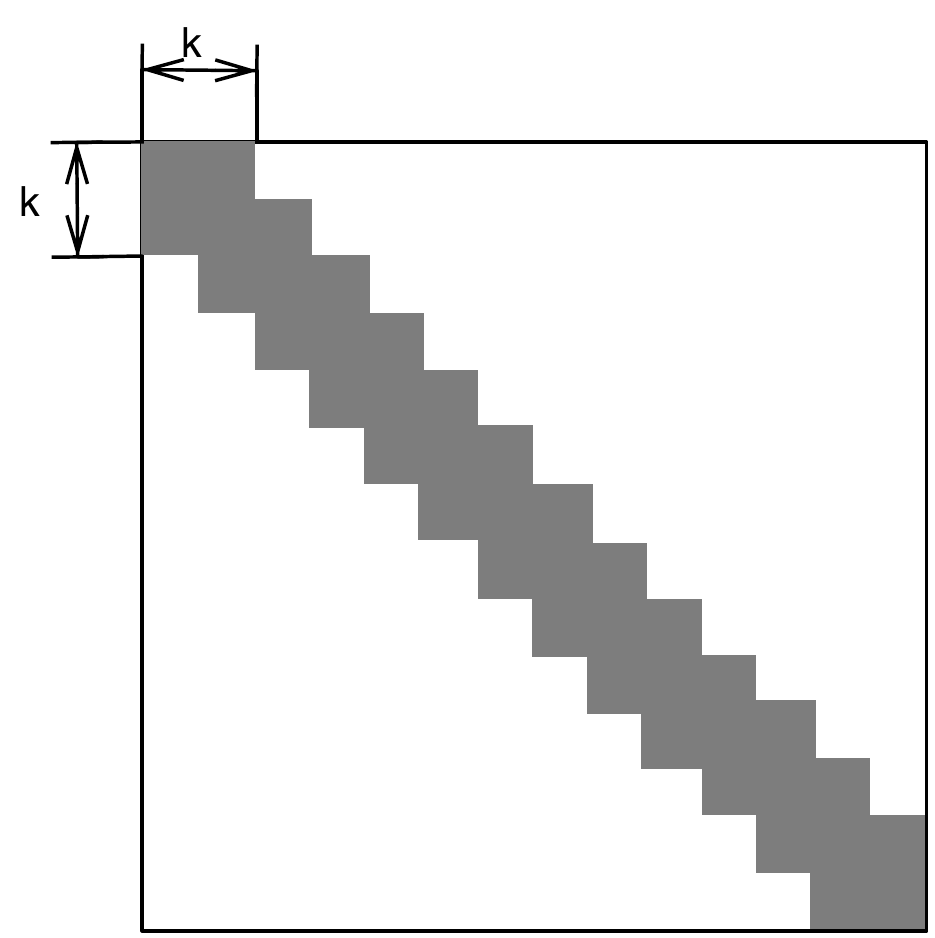}}
\caption{In a) we show the graph of a 3-BAND model for six different variables. In b) we show the connectivity matrix for a general K-BAND model.}
\end{figure}

% 
% \begin{figure}
% \centering
% \includegraphics[width=0.2\textwidth]{Figures/cov-k-band.eps}
% \caption{Structure of the covariance matrix for the $k$-BAND model.}
% 
% \end{figure}
%\begin{figure}
%\centering
%\subfigure[$k$-BOX model]{\label{fig:kboxgraph}\includegraphics[width=0.3\textwidth]{Figures/k-box-graf.eps}}\\
%\subfigure[$k$-BAND model]{\label{fig:kbandgraph}\includegraphics[width=0.3\textwidth]{Figures/k-band-graf.eps}}\\
%\caption{Structure of the graph for different models}
%\end{figure}
% \end{figure}
% \begin{figure}
% \centering
% \includegraphics[width=0.35\textwidth]{Figures/k-band-graf.eps}
% \caption{Structure of the graph for the $k$-BAND model with $k=3$.}
% \label{fig:kbandgraph}
% \end{figure}
The second structure proposed is the $k$-BAND structure. In $k$-BAND, we assume that each variable is independent of all the remaining variables given the $k-1$ variables that precede it and the class variable. 
%force a variable $X_i$ to be the parent of the next $k-1$ variables in the graph. 
%Therefore, the $k$-BAND structure is formed by cliques of size $k$ with separators of $k-1$ variables.%, i.e., given two cliques $C_i = \{X_i,X_{i+1},\ldots,X_{i+k-1}\}$ and $C_{i+1} = \{X_{i+1},\ldots,X_{i+k+2}\}$ then the separator of this two cliques is $S_i = \{X_{i+1},\ldots,X_{i+k+1}\}$.

The covariance matrix for a $k$-BAND structure is a band of size $k$ around the diagonal, as is shown in Figure \ref{fig:kbandmat}. An example of the structure is shown in Figure \ref{fig:kbandgraph}.
% a  classifier with 6 predictor variables with $k=3$  were we have 4 different cliques $C_1=\{X_1,X_2,X_3\}$, $C_2=\{X_2,X_3,X_4\}$, $C_1=\{X_3,X_4,X_5\}$ and $C_4=\{X_4,X_5,X_6\}$. 

We ran a sequence of experiments  to compare the different structures ($k$-BOX and $k$-BAND) and parameter learning methods (BA and ML) varying  the $k$ parameter from 1 to 50. 
%We have run it for different sizes of the cliques for each structure. The parameters of the structures are learned using ML (ML) and Bayesian model averaging (BMA) over the parameters.  

We performed 5 repetitions of 10-fold cross validation and assessed the accuracy and CLL.

\begin{figure}
\hspace{-0.5cm}
\subfigure[Accuracy]{\label{fig:ovAcc}\includegraphics[width=0.55\textwidth]{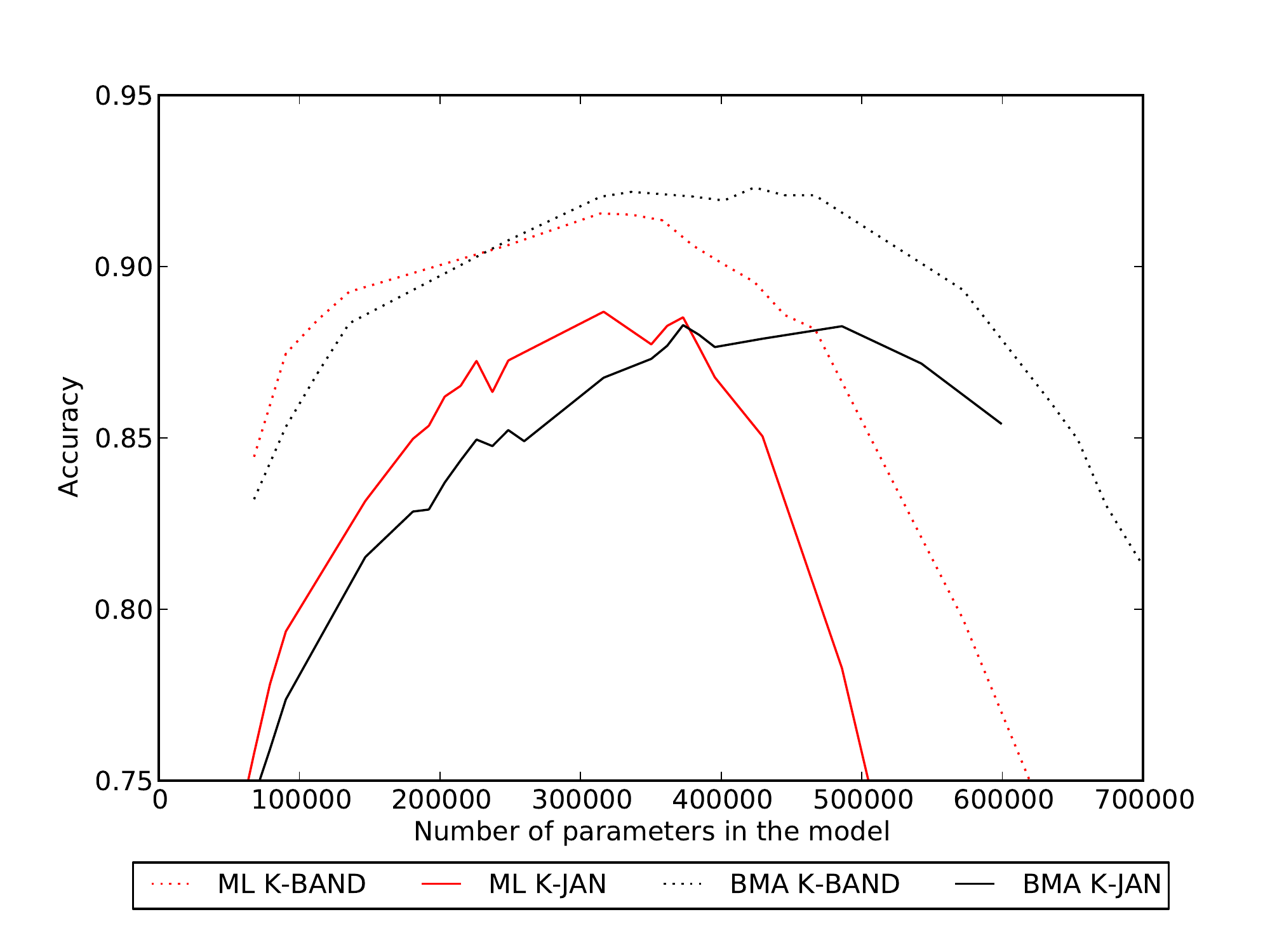}}\hspace{-0.5cm}
\subfigure[CLL]{\label{fig:ovCLL}\includegraphics[width=0.55\textwidth]{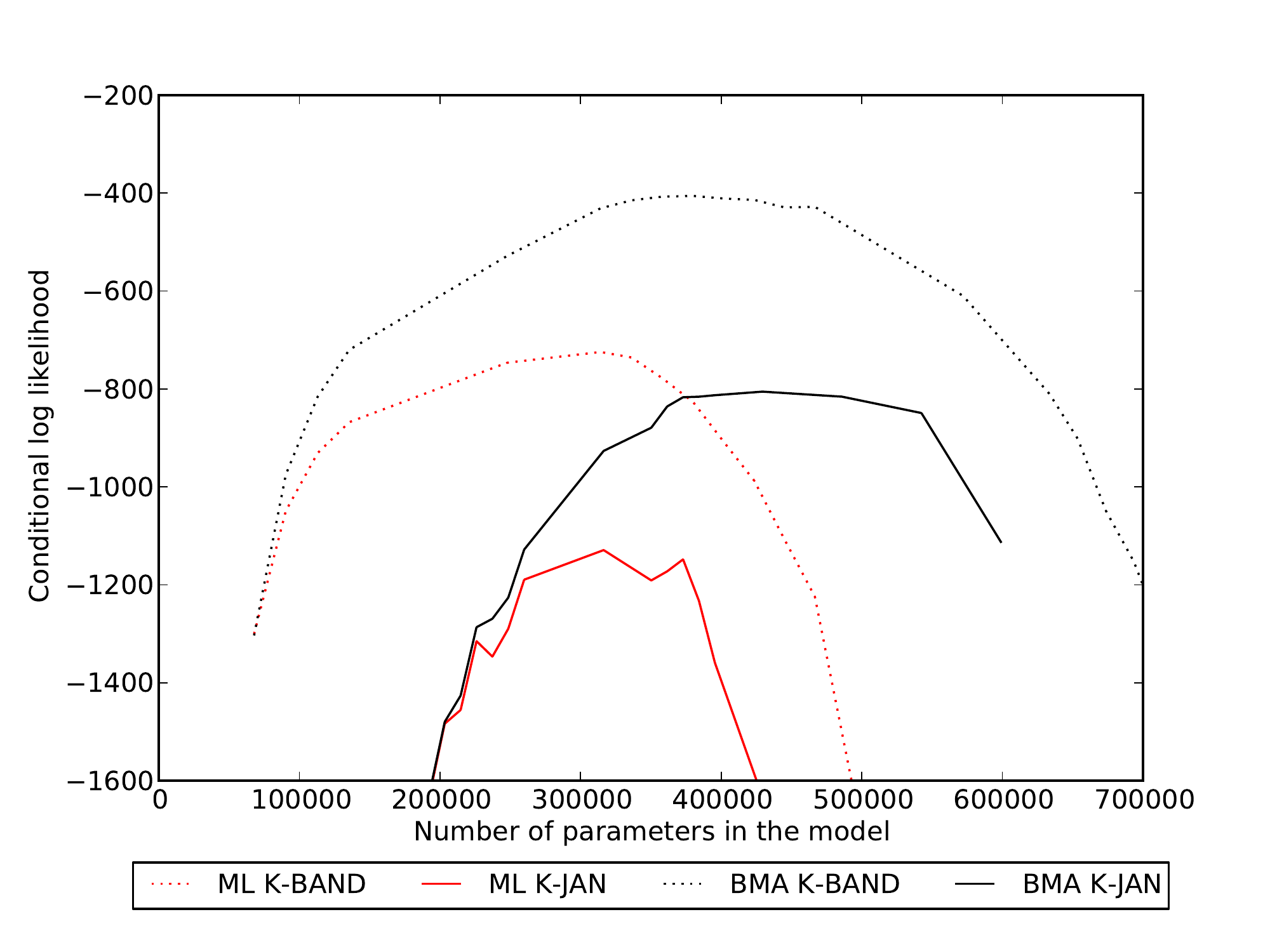}}
  \caption{Prediction of ovarian cancer.In the horizontal axis we present the number of parameters in model. In the vertical axis we show the accuracy and the CLL.}

\end{figure}

Figure \ref{fig:ovAcc} shows the mean accuracy versus the number of parameters in the model. We see that $k$-BAND models are more accurate than $k$-BOX models. Furthermore, $k$-BAND models learned using BA outperform those learned using ML when the number of parameters is large.

Figure \ref{fig:ovCLL} shows the mean CLL versus the number of parameters in each structure. Again,  $k$-BAND models outperform $k$-BOX models. Furthermore for both $k$-BOX and $k$-BAND models, using BA significantly increases the quality of the probabilities predicted. 
%For low values of $k$, ML has a  but GJT has a largest accuracy and a highest CLL at its peak, and shows a more graceful decay as the number of parameters grows beyond that peak.  

%The pancreatic cancer data has been classified using $k$-BOX and $k$-BAND structures with $k$ ranging from 1 to 30. Figure~\ref{fig:panAcc} and  Figure~\ref{fig:panCLL} show respectively the mean accuracy and mean CLL against the number of parameters. Note that the accuracy for this dataset is much lower and close to the frequency of the largest class, that is $55.8\%$ in this datasets. Previous studies have shown that the accuracy results for this dataset are much lower than for the previous one. The $k$-BOX model using GJT appears to reach the highest accuracy and the $k$ -BAND model reaches the highest CLL also for pancreatic cancer. 

% \begin{figure}
%   \centering
%     \includegraphics[width=0.47\textwidth]{{Figures/pancreas-accuracy-nv}.eps}
%   \caption{Prediction of pancreatic cancer. Accuracy versus number of parameters in model. }
%   \label{fig:panAcc}
% \end{figure}
% \begin{figure}
%   \centering
%     \includegraphics[width=0.47\textwidth]{{Figures/pancreas-cll-nv}.eps}
%   \caption{Prediction of pancreatic cancer. CLL versus number of parameters in model.}
%   \label{fig:panCLL}
% \end{figure}

\section{Conclusions and future work}
\label{sec:Conclusions}
We have analyzed two alternatives for dealing with parameters in CGNs: ML and BA. Our experiments confirm that BA results in a classifier that estimates better the probabilities of the different classes. Furthermore, we have seen that this effect shows up more clearly as the number of variables is large with respect to the number of instances. Since this situation is common in areas such as bioinformatics, we have provided an example application of this approach to the problem of diagnosing ovarian cancer from mass sprectra. Finally, an open source implementation of the algorithms described in the paper is provided for free use at \url{http://www.iiia.csic.es/~cerquide/pypermarkov}.

In this work, we have focused on learning CGN classifiers from a generative approach. Directly maximizing the CLL following a discriminative approach is a future line of research, as it is the study of priors for Bayesian linear regression other than the $\mathcal{NIG}$.

\appendix
\newpage
\section{Distributions}
\label{sec:Distributions}
\begin{definition}[Multivariate normal]
We say that $\boldsymbol{\beta}\in \mathbb{R}^p,$ follows a multivariate normal distribution with parameters $\boldsymbol{\mu} \in \mathbb{R}^p,V \in \mathbb{R}^p\times\mathbb{R}^p$,  $V$ symmetric and positive definite if
\begin{equation}
 p(\boldsymbol{\beta}) = \mathcal{N}(\boldsymbol{\beta}|\boldsymbol{\mu}, V) = \frac{1}{(2\pi) ^{p/2}|V|^{1/2}}\exp\left(-\frac{1}{2}(\boldsymbol{\beta}-\boldsymbol{\mu})^T V^{-1} (\boldsymbol{\beta}-\boldsymbol{\mu}) \right)
\end{equation}
\end{definition}

\begin{definition}[Inverse Gamma]
We say that $\sigma^2\in \mathbb{R}$ follows an inverse gamma distribution with parameters $\rho>0,\phi>0$ if
\begin{equation}
 p(\sigma^2) = \mathcal{IG}(\sigma^2|\rho,\phi) = \frac{\phi^\rho}{\Gamma(\rho)}\left(\frac{1}{\sigma^2}\right)^{\rho+1} \exp\left(-\frac{\phi}{\sigma^2}\right)
\end{equation}
\end{definition}

\begin{definition}[Normal Inverse Gamma]
We say that $\boldsymbol{\beta}\in \mathbb{R}^p,$ and $\sigma^2\in \mathbb{R}$ follow a normal inverse gamma distribution with parameters $\boldsymbol{\mu} \in \mathbb{R}^p,V \in \mathbb{R}^p\times\mathbb{R}^p,\rho>0,\phi>0$,  $V$ symmetric and positive definite if
\begin{equation}
 p(\boldsymbol{\beta}, \sigma^2) = \mathcal{NIG}(\boldsymbol{\beta}, \sigma^2|\boldsymbol{\mu},V,\rho,\phi) = \mathcal{IG}(\sigma^2|\rho,\phi) \times \mathcal{N}(\boldsymbol{\beta}|\boldsymbol{\mu},\sigma^2 V)
\end{equation}
\end{definition}

\begin{definition}[Multivariate Student]
We say that $\mathbf{x}\in \mathbb{R}^m,$ follows multivariate Student distribution with parameters $\nu>0,\boldsymbol{\mu}\in \mathbb{R}^m,\Sigma\in \mathbb{R}^m\times \mathbb{R}^m$ 
with $\Sigma$ symmetric and positive definite if
\begin{multline}
 p(\mathbf{x}) = MVSt(\mathbf{x}|\nu,\boldsymbol{\mu},\Sigma) = \\
 \frac{\Gamma((\nu+m)/2)}{(\pi \nu)^{m/2}\Gamma(\nu/2)|\Sigma|^{1/2}} \left[1+\frac{1}{\nu}(\mathbf{x} - \boldsymbol{\mu})^T \Sigma^{-1}(\mathbf{x}-\boldsymbol{\mu})\right]^{-(\nu+m)/2}
\end{multline}
\end{definition}

\begin{definition}[Multinomial]
We say that an $m$-valued discrete random variable $X$ taking values in set $\mathcal X=\langle x_1,\ldots,x_m\rangle$ follows a multinomial distribution with parameters $\boldsymbol{\psi}=\langle\psi_{x_1},\ldots, \psi_{x_m}\rangle \in \mathbb{R}^m$ with $\psi_{x}>0$  $\forall x\in \mathcal X$ and $\sum_{x \in \mathcal X}\psi_{x}=1$ if
\begin{equation}
 p(x|\boldsymbol{\psi})=\mathcal{MN}(x|\boldsymbol{\psi}) = \psi_{x}\ \ \ \ \ \ \forall x\in \mathcal X
\end{equation}
\end{definition}

\begin{definition}[Dirichlet]
We say that $\boldsymbol{\theta}\in \mathbb{R}^m,$ follows a Dirichlet distribution with parameters $\boldsymbol{\psi} \in \mathbb{R}^m$ if with $\psi_i>0$ for all $1<i\leq m$ if
\begin{equation}
 p(\boldsymbol{\theta})=\mathcal{D}(\boldsymbol{\theta}|\boldsymbol{\psi}) = \frac{\prod_{i=1}^m \Gamma(\psi_i)}{\Gamma\left(\sum_{i=1}^m \psi_i\right)}  \prod_{i=1}^m \theta_i^{\psi_i}
\end{equation}
\end{definition}

\section{Bayesian multinomial model}
\label{sec:BayesianMultinomial}
Let $X$ be a discrete random variable, with domain $\mathcal{X}$ having $m$ different values, following a multinomial distribution.
\begin{equation}
p(x|\boldsymbol{\theta}) = \mathcal{MN}(x|\boldsymbol{\theta}).
\end{equation}

If we are uncertain about the values of $\boldsymbol{\theta}\in \mathbb{R}^m$, we are given $\mathbf{x}=[x_i]^n_{i=1}$ a $n\times 1$ vector of independent observations from $n$ experimental units, and we assume as prior $p(\boldsymbol{\theta}|\boldsymbol{\Psi})=\mathcal{D}(\boldsymbol{\theta}|\boldsymbol{\Psi})$, the posterior after observing $\mathbf{x}$ is a Dirichlet distribution $p(\boldsymbol{\theta}|\boldsymbol{\Psi})=\mathcal{D}(\boldsymbol{\theta}|\boldsymbol{\Psi'})$ with
\begin{equation}
\Psi'_w = \Psi_w + n(\mathbf{x}=w) \ \ \ \ \ \forall w\in \mathcal{X} 
\end{equation}
where $n(\mathbf{x}=w)$ is the number of times that value $w$ is observed in the sample.

Provided $p(\boldsymbol{\theta}|\boldsymbol{\Psi}) = \mathcal{D}(\boldsymbol{\theta}|\boldsymbol{\Psi})$ we have that $p(x|\boldsymbol{\Psi}) = \mathcal{MN}(x|\boldsymbol{\boldsymbol{\varphi}})$ with
\begin{equation}
\varphi_x = \frac{\Psi_x}{\sum_{x'\in\mathcal{X}}{\Psi_{x'}}}  \ \ \ \ \ \ \ \forall x\in \mathcal{X}.
\end{equation}

\section{Bayesian linear regression}
\label{sec:BayesianGaussianLinearRegression}
In this section we summarize the Bayesian linear results needed in the paper. The main ideas come from the seminal paper of Lindley and Smith \cite{Lindley1972}. The results are provided here for easy reference. Proofs and intuitive explanations can be found in chapter 3 of \cite{Koop2003} and in \cite{Banerjee2012}. 
\subsection{The Gaussian linear regression model}
Let $Y, X_1, \cdots, X_p$ be continuous random variables. The Gaussian linear regression model assumes that
\begin{equation}
p(y|\boldsymbol{\beta},\sigma^2) = {\mathcal N}(y|\boldsymbol{\beta}^T\mathbf{x},\sigma^2).
\end{equation}
where $y$ is the observation of the dependent variable, $\boldsymbol{\beta}$ is the $p \times 1$ slope vector of regression coefficients, $\mathbf{x}$ is the $p \times 1$ vector of regressors and $\sigma^2 \in \mathbb{R}$ is the variance.  

\subsection{Bayesian learning with the Gaussian linear regression model}
Assume we are uncertain about the values of $\boldsymbol{\beta}$ and $\sigma^2$. Furthermore, say that we are given  $\mathbf{y}=[y_i]^n_{i=1}$ a $n\times 1$ vector of independent observations on the dependent variable (or response) from $n$ experimental units. Associated with each $y_i$, is a $p \times 1$ vector of regressors, say $\mathbf{x}_i$. Furthermore $X = [\mathbf{x}_i]^n_{i=1}$ is the $p \times n$ matrix of regressors with $i$-th column being $\mathbf{x}_i.$  We are expected to improve our knowledge about $\boldsymbol{\beta}$ and $\sigma^2$ from $\mathbf{y}$ and $X.$ First, we need to represent our initial uncertain knowledge as a probability distribution over $\boldsymbol{\beta}$ and $\sigma^2$. In this case the conjugate distribution is the normal inverse gamma. Thus, we assume as prior $\xi=(\boldsymbol{\mu},V,\rho,\phi)$ that
\begin{equation}
p(\boldsymbol{\beta},\sigma^2|\xi) =  \mathcal{NIG}(\boldsymbol{\beta}, \sigma^2|\boldsymbol{\mu},V,\rho,\phi) 
 \end{equation}
The posterior after observing $\mathbf{y},X$ is 
\begin{equation}
p(\boldsymbol{\beta},\sigma^2|\xi,\mathbf{y},X) =  \mathcal{NIG}(\boldsymbol{\beta}, \sigma^2|\boldsymbol{\mu}',V',\rho',\phi') 
\end{equation}
where 
\begin{alignat}{4}
&V' & = &\  (V^{-1} + X^\top X) ^{-1},\\
&\boldsymbol{\mu}' \ & = & \ V'(V^{-1}\boldsymbol{\mu} + X^T\mathbf{y}),\\
&\rho' &= &\ \rho + n/2,\\
&\phi' &=&\  \phi + \frac{1}{2}[\boldsymbol{\mu}^T V^{-1}\boldsymbol{\mu} + \mathbf{y}^T\mathbf{y} - {\boldsymbol{\mu}'}^\top {V'}^{-1} \boldsymbol{\mu'}] 
\end{alignat}

\subsection{Bayesian prediction with the Gaussian linear regression model}
Let $\mathbf{\tilde{y}}=[\tilde{y}_i]^m_{i=1}$ be an unknown $m\times 1$ vector of independent observations on the dependent variable from $m$ new experimental units and $\tilde{X}$ the corresponding observed matrix of regressors. If $\boldsymbol{\beta},\sigma^2$ follow a normal inverse gamma with parameters $\boldsymbol{\mu},V,\rho,\phi$, the probability distribution for $\tilde{\mathbf{y}}$ given $\tilde{X}$ is
\begin{equation}
 p(\mathbf{\tilde{y}}|\tilde{X}) = MVSt\left(\mathbf{\tilde{y}}|2\rho,\tilde{X}\boldsymbol{\mu},\frac{\phi}{\rho}(I+\tilde{X}V\tilde{X}^T)\right)
\end{equation}

\bibliographystyle{elsarticle-num}
\bibliography{library}
\end{document}